# PETWB-REP: A Multi-Cancer Whole-Body FDG PET/CT and Radiology Report Dataset for Medical Imaging Research

Le Xue[1,2†*], Gang Feng[3,4†], Wenbo Zhang[2,5], Yichi Zhang[2,6], Lanlan Li[5], Shuqi Wang[5], Liling Peng[4], Sisi Peng[4*], Xin Gao[4*]

1 PET Center, Huashan Hospital, Fudan University, Shanghai, China.
2 Shanghai Academy of Artificial Intelligence for Science, Shanghai, China.
3 Institute of Science and Technology for Brain-inspired Intelligence, Fudan University, Shanghai, China
4 Shanghai Universal Medical Imaging Diagnostic Center, Shanghai, China
5 Human Phenome Institute, Fudan University, Shanghai, China.
6 School of Data Science, Fudan University, Shanghai, China.

\* Corresponding author(s): Xin Gao (gaoxin@uvclinic.cn); Sisi Peng (pengsisi@uvclinic.cn); Le Xue (lexue@fudan.edu.cn)

† These authors contributed equally to this work

**Abstract**

Publicly available, large-scale medical imaging datasets are crucial for developing and validating artificial intelligence models and conducting retrospective clinical research. However, datasets that combine functional and anatomical imaging with detailed clinical reports across multiple cancer types remain scarce. Here, we present PETWB-REP, a curated dataset comprising whole-body [18]F-Fluorodeoxyglucose (FDG) Positron Emission Tomography/Computed Tomography (PET/CT) scans and corresponding radiology reports from 490 patients diagnosed with various malignancies. The dataset primarily includes common cancers such as lung cancer, liver cancer, breast cancer, prostate cancer, and ovarian cancer. This dataset includes paired PET and CT images, de-identified textual reports, and structured clinical metadata. It is designed to support research in medical imaging, radiomics, artificial intelligence, and multi-modal learning.

**Background & Summary**

$^{18}$F-FDG PET/CT remains a mainstay in cancer imaging for staging, monitoring, and treatment response evaluation[1,2]. The fusion of functional metabolic information from PET with detailed anatomical information from CT provides a comprehensive view of tumor burden and activity. While numerous studies have demonstrated the utility of PET-CT[3-5], the public availability of large-scale, well-curated datasets remains limited. Existing resources often focus on a single cancer type, may lack corresponding clinical reports, or have limited sample sizes, which restricts the development of generalizable and robust computational models[6-8].

The rapid advancement of artificial intelligence (AI), particularly deep learning, has demonstrated the potential to transform the analysis of these complex medical images[9-11]. AI models are being developed to automate a range of tasks, from lesion detection and segmentation to the prediction of clinical outcomes[12-16]. However, the efficacy and generalizability of these models are fundamentally contingent upon the availability of large-scale, high-quality, and well-curated datasets.

To address this gap, we constructed PETWB-REP, a dataset containing 490 PET/CT studies from a single medical imaging center, encompassing a diverse cancer population. All cases include de-identified imaging data, fully anonymized radiology reports, and key clinical variables. This dataset enables a wide range of downstream research applications, including radiomic biomarker discovery, automated report generation, tumor segmentation, and multi-modal fusion learning.

**Methods**

**Subject characteristics**

The data were retrospectively collected from patients who underwent clinically indicated whole-body $^{18}$F-FDG PET/CT scans at the Shanghai Universal Medical Imaging Diagnostic Center between 2021 and 2024. The requirement for individual patient consent was waived by the IRB due to the retrospective nature of the study and the full de-identification of the data.

Inclusion criteria were: (1) patients with a confirmed diagnosis of malignancy; (2) availability of whole-body FDG PET/CT scan data; (3) availability of the corresponding clinical radiology report. Exclusion criteria were: (1) significant image artifacts (e.g., motion, metal) that would preclude analysis; (2) incomplete or corrupted image data. A total of 490 patients ( 219 females and 271 males) were recruited. The mean age of the subjects was 60.98 ±12.77 years. The distribution of cancer types in the study is illustrated in Figure. 1. Lung cancer was the most prevalent (34.29%), followed by liver cancer (10.00%) and cervical cancer (7.76%). Other cancer types each accounted for less than 6% of cases, including pancreatic cancer (5.51%), lymphoma (5.31%), and renal cancer (5.31%). The remaining cancer types collectively contributed to the minority of cases.

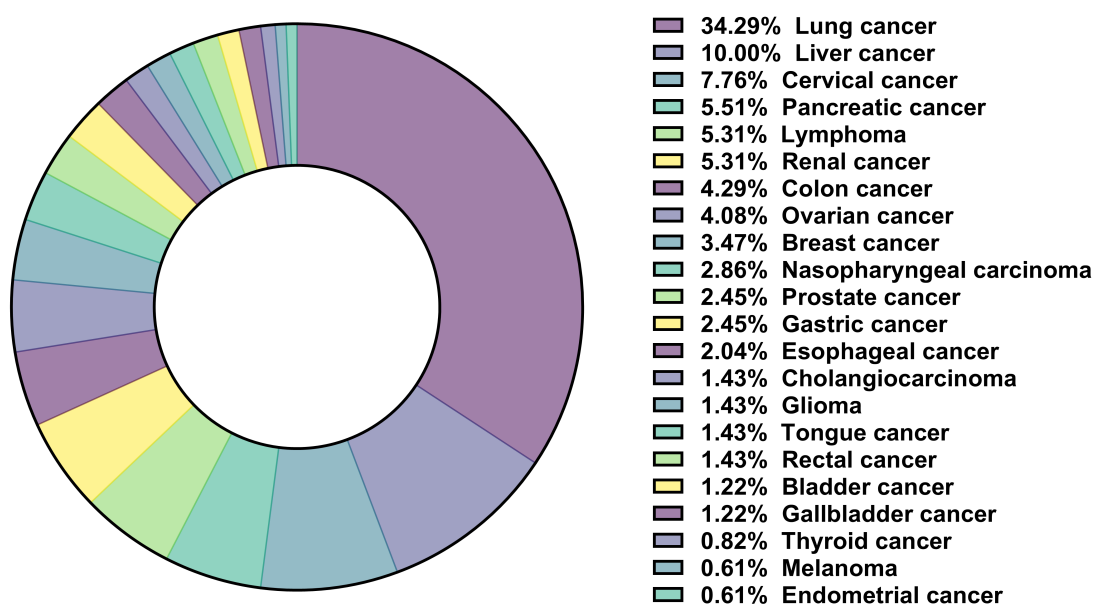

**Figure 1.** The distribution of cancer types in the study.

**Image acquisition protocol**

All patients fasted for at least 6 hours prior to scanning. Blood glucose levels were measured before tracer administration and confirmed to be below 11.1 mmol/L. Each patient received an intravenous injection of $^{18}$F-FDG at a dose of 3.70-5.55 MBq/kg. Following tracer injection, patients rested in a quiet, dimly lit room for a 60-minute uptake period. Imaging was performed using a Biograph 64 PET/CT scanner (Siemens Healthcare, Germany). The scanning range extended from the base of the skull to the mid-thigh, with patients positioned supine. A low-dose CT scan was first acquired for attenuation correction and anatomical localization, using the following parameters: 120 kV tube voltage, 170 mA tube current, and 3.0 mm slice thickness. Subsequently, a PET scan was acquired in 3D mode, covering 5-6 bed positions with an acquisition time of 2.5 minutes per bed. PET images were reconstructed using the Ordered Subsets

Expectation Maximization (OSEM) algorithm with 2 iterations and 21 subsets, followed by post-reconstruction smoothing with a 5-mm Gaussian filter.

**Radiology report processing and de-identification**

The radiology reports were authored by a board-certified nuclear medicine physician with over 10 years of clinical experience, and independently reviewed by a senior nuclear medicine specialist with more than 20 years of diagnostic expertise. Findings are systematically organized by anatomic region (e.g., head and neck, chest, abdomen and pelvis, musculoskeletal), with both PET and CT findings described within each respective subsection. Nodules and masses are reported with a single transaxial diameter, and lesion size is explicitly documented. The metabolic activity of lesions is quantified using the maximum standardized uptake value (SUVmax), providing a semiquantitative measure of $^{18}$F-FDG uptake. The Impression section offers a concise summary and clinical interpretation of the key imaging findings, highlighting areas of diagnostic relevance. These reports were exported from the Picture Archiving and Communication System (PACS) in text format.

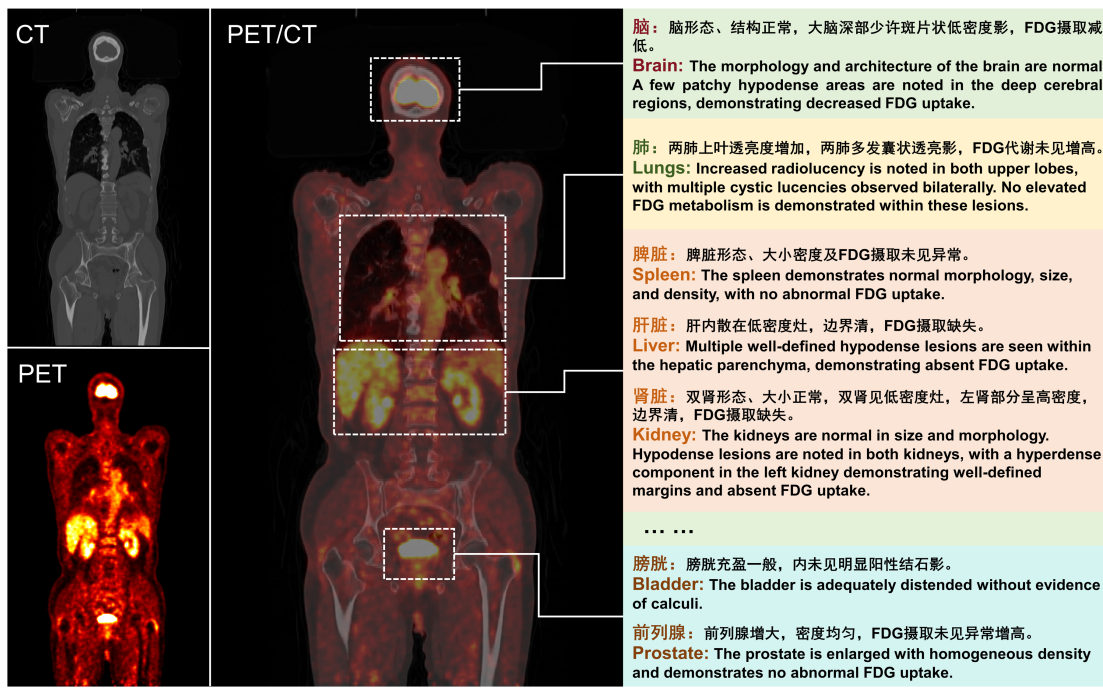

**Figure 2.** Representative whole-body FDG PET/CT images along with the corresponding radiology report.

The original radiology reports were authored in Chinese by board-certified physicians as part of routine clinical practice. To make the data accessible to the international research community, all reports were translated into English. The translation process followed a rigorous two-step protocol to ensure quality and accuracy. First, each report was initially translated from Chinese to English using Google Translator. Second, every translated report was reviewed, edited, and validated by a senior nuclear medicine physician (S. Peng) fluent in both Chinese and English to ensure that all medical terminology was translated correctly and the clinical meaning was preserved. For data transparency and validation purposes, both the original Chinese version (*_zh.csv) and the validated English version (*_en.csv) of each report are provided in the dataset. Figure. 2 illustrates representative whole-body FDG PET/CT images along with the corresponding radiology report.

**Data preprocessing**

An overview of the data generation pipeline is presented in Figure. 3.

**Anonymization:** To protect patient privacy, a rigorous de-identification process was applied to both the DICOM image headers and the text reports. All Protected Health Information (PHI), including but not limited to patient names, ID numbers, accession numbers, dates of birth, and specific dates of procedures, was removed or replaced with generic, non-identifiable codes. Then, two researchers manually reviewed all files to ensure complete de-identification. Each patient was assigned a unique, random identifier which is consistent across their imaging files and report.

**Format Conversion:** The anonymized DICOM series for PET and CT were converted to the Neuroimaging Informatics Technology Initiative (NIfTI) format (.nii.gz) using the dcm2niix tool.

**Image Normalization:** Raw CT images were normalized using z-score standardization to enhance grayscale contrast and improve downstream processing. For PET images, raw counts were converted into Standardized Uptake Values normalized by body weight (SUVbw). The calculation was performed using patient weight and injected dose information extracted from the DICOM metadata, following the formula:

$$SUV_{bw} = \frac{RC(kBq/mL)}{ID(MBq)/BW(kg)}$$

where RC represents the tissue radioactivity concentration, ID is the injected dose, and BW is the body weight. The resulting SUV maps were saved as NIfTI format for further analysis.

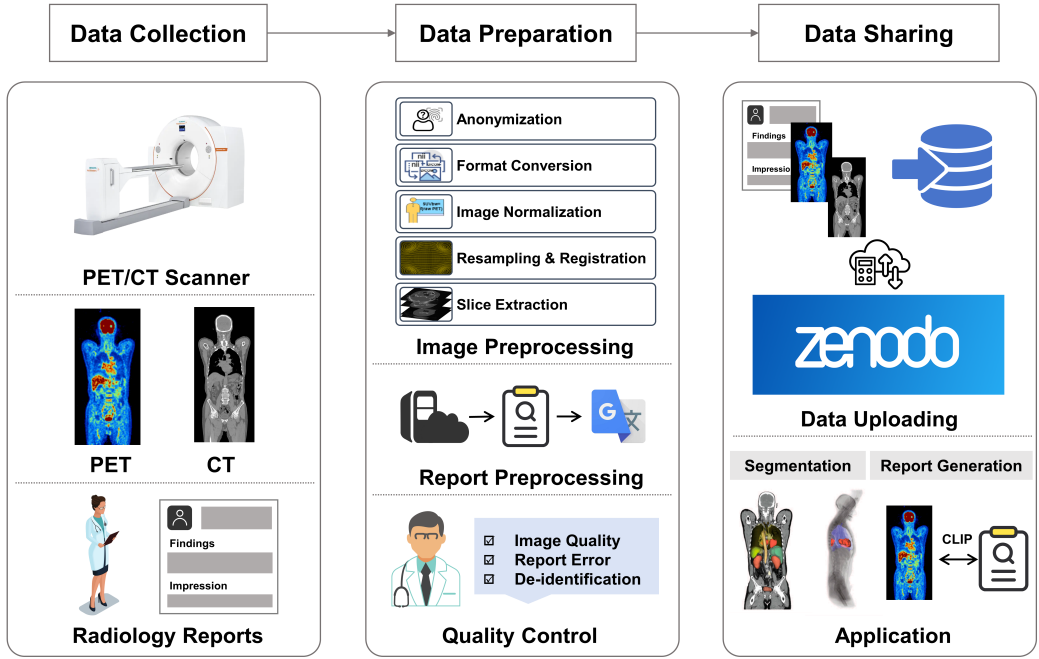

**Figure 3.** An overview of the data generation and processing pipeline for the PETWB-REP dataset. The workflow begins with the collection of PET/CT images and corresponding radiology reports. The data then undergoes a comprehensive preparation phase that includes anonymization to remove all Protected Health Information (PHI), format conversion to NIfTI, image normalization, resampling and registration, and report preprocessing. A rigorous quality control process follows, which involves checks for image quality, data integrity, and successful de-identification. The finalized dataset is then uploaded and shared via the Zenodo repository, ready for use in various downstream applications.

**Resampling and Registration:** CT and SUV-PET images were resampled to an isotropic in-plane resolution of 0.98×0.98 mm and a slice thickness of 3.00 mm using transformation matrices for spatial alignment. Subsequently, PET images were registered to the CT space using B-spline interpolation, which captures local anatomical deformations while preserving overall image fidelity. Both the original and resampled image volumes were provided.

**Slice Extraction:** Axial slices were extracted from both the original and processed CT and PET image volumes. Extraction was performed sequentially from the cranial to the caudal direction to ensure consistent spatial correspondence. The original image resolution and orientation were preserved during slicing to maintain anatomical accuracy and data integrity.

**Metadata Compilation:** A master metadata file (metadata.csv) was created. Information was extracted from anonymized DICOM headers (e.g., age, sex), de-identified clinical records (e.g., cancer type), and the imaging protocol logs (e.g., uptake time, tracer dose).

**Data Records**

The directory structure is shown in Figure. 4. The dataset is organized into two main folders: Imaging_data and Non-Imaging_data. The Imaging_data folder contains raw and processed CT and PET images for each subject (e.g., sub_001), including raw CT images (CT_RAW), raw PET images (PET_RAW), z-score normalized CT images (CT_NORM), and SUV-converted PET images (PET_SUV). The Non-Imaging_data folder stores associated metadata (meta_data.csv) and de-identified radiology reports in both English (report_en.csv) and the original Chinese (report_zh.csv).

**Technical Validation**

To ensure the quality and usability of the dataset, several validation steps were performed.

**Image Quality Control:** All cases were visually inspected by an experienced nuclear medicine physician (L. Peng) to check for the presence of major artifacts (e.g., motion, truncation, metal artifacts) and to confirm proper image registration between PET and CT.

**De-identification Confirmation:** The de-identification process was validated by having a separate researcher (S. Wang), not involved in the initial anonymization, review a of NIfTI headers and reports to confirm the absence of any remaining PHI.

**Data Integrity:** We verified that for every patient folder, the corresponding image series (PET and CT) and the report file are present as listed in the metadata.csv file.

**Report-Metadata Consistency:** The primary diagnosis listed in metadata.csv was cross-checked against the content of the radiology report for all cases to ensure accuracy.

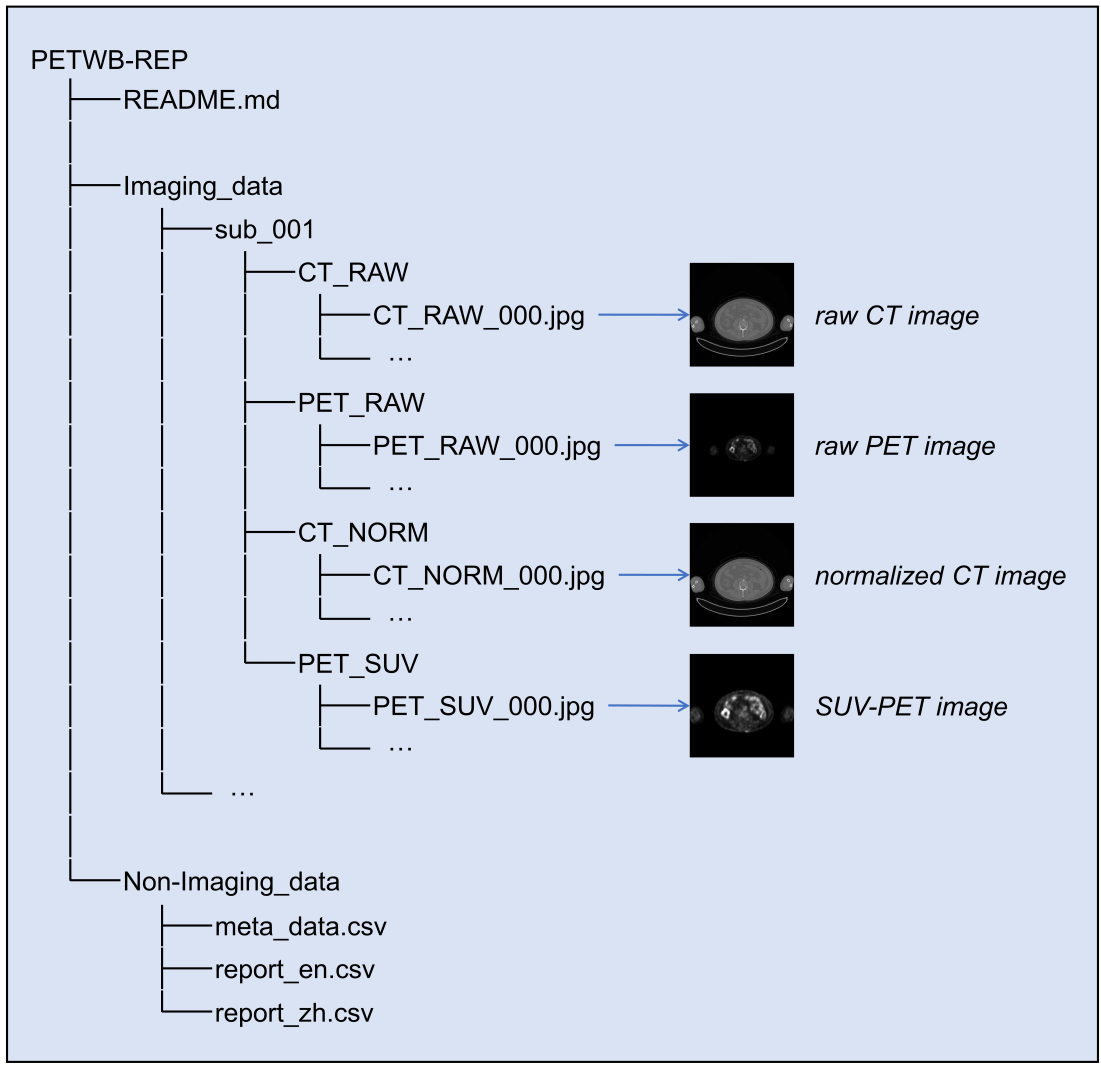

**Figure 4.** The directory structure of the PETWB-REP dataset.

**Usage Notes**

The PETWB-REP dataset is suitable for a variety of image analysis tasks, including but not limited to:

1. Deep learning model training for tumor detection or segmentation: Training, validating, and testing models for automated tumor detection, segmentation, and classification. The multi-organ, multi-cancer nature of the data is suitable for developing generalizable models.

2. Natural Language Processing (NLP): Developing and testing NLP models to automatically extract structured information (e.g., lesion locations, measurements) from free-text radiology reports.

3. Multi-modal Fusion: Investigating methods that combine information from imaging (PET/CT) and text reports to improve diagnostic or prognostic accuracy.

**Limitations:** This dataset is derived from a single institution, which may introduce biases related to scanner type, patient demographics, and local reporting practices. Additionally, the retrospective nature of data collection means that certain clinical variables may be missing or inconsistently recorded. The distribution of cancer types is non-uniform and reflects the clinical referral pattern of the imaging center.

Despite these limitations, this large, multi-modal, and multi-cancer dataset represents a valuable resource for the medical imaging and oncology research communities.